\documentclass[conference]{IEEEtran}
\IEEEoverridecommandlockouts
\usepackage{graphicx}
\usepackage{cite}
\usepackage{amsmath,amssymb,amsfonts}
\usepackage{algorithmic}
\usepackage{physics}
\usepackage{textcomp}
\usepackage{subcaption}
\usepackage[dvipsnames]{xcolor}
\usepackage{tabularx}
\usepackage[utf8x]{inputenc} 

\begin{document}

\title{
\begin{center}
{\LARGE \bf Modularis: Modular Underwater Robot for\\[6pt]
Rapid Development and Validation of Autonomous Systems} \\[8pt]
\end{center}
}

\author{\IEEEauthorblockN{Baker Herrin}
\IEEEauthorblockA{\textit{Mechanical and Aerospace Engineering} \\
\textit{University of Florida}\\
Gainesville, FL, United States \\
eherrin@ufl.edu}
\and
\IEEEauthorblockN{Victoria Close}
\IEEEauthorblockA{\textit{Electrical and Computer Engineering} \\
\textit{University of Florida}\\
Gainesville, FL, United States \\
vclose@ufl.edu}
\and
\IEEEauthorblockN{Nathan Berner}
\IEEEauthorblockA{\textit{Mechanical and Aerospace Engineering} \\
\textit{University of Florida}\\
Gainesville, FL, United States \\
nberner@ufl.edu}
\and
\IEEEauthorblockN{Joshua Hebert}
\IEEEauthorblockA{\textit{Mechanical and Aerospace Engineering} \\
\textit{University of Florida}\\
Gainesville, FL, United States \\
j.hebert@ufl.edu}
\and
\IEEEauthorblockN{Ethan Reussow}
\IEEEauthorblockA{\textit{Electrical and Computer Engineering} \\
\textit{University of Florida}\\
Gainesville, FL, United States \\
ryan.james@ufl.edu}
\and
\IEEEauthorblockN{Ryan James}
\IEEEauthorblockA{\textit{Electrical and Computer Engineering} \\
\textit{University of Florida}\\
Gainesville, FL, United States \\
ryan.james@ufl.edu}
\and
\IEEEauthorblockN{Cale Woodward}
\IEEEauthorblockA{\textit{Electrical and Computer Engineering} \\
\textit{University of Florida}\\
Gainesville, FL, United States \\
calewoodward@ufl.edu}
\and
\IEEEauthorblockN{Jared Mindlin}
\IEEEauthorblockA{\textit{Electrical and Computer Engineering} \\
\textit{University of Florida}\\
Gainesville, FL, United States \\
jared.mindlin@ufl.edu}
\and
\IEEEauthorblockN{Sebastian Paez}
\IEEEauthorblockA{\textit{Electrical and Computer Engineering} \\
\textit{University of Florida}\\
Gainesville, FL, United States \\
carlos.paez@ufl.edu}
\and
\IEEEauthorblockN{Nilson Bretas}
\IEEEauthorblockA{\textit{Electrical and Computer Engineering} \\
\textit{University of Florida}\\
Gainesville, FL, United States \\
nbretas@ufl.edu}
\and
\IEEEauthorblockN{      }
\IEEEauthorblockA{\textit{              } \\
\textit{}\\
 \\
}
\and
\IEEEauthorblockN{Jane Shin}
\IEEEauthorblockA{\textit{Mechanical and Aerospace Engineering} \\
\textit{University of Florida}\\
Gainesville, FL, United States \\
jane.shin@ufl.edu}
}

\maketitle

\begin{abstract}
Autonomous underwater robots typically require higher cost and time for demonstrations compared to other domains due to the complexity of the environment. Due to the limited capacity and payload flexibility, it is challenging to find off-the-shelf underwater robots that are affordable, customizable, and subject to environmental variability. Custom-built underwater robots may be necessary for specialized applications or missions, but the process can be more costly and time-consuming than purchasing an off-the-shelf autonomous underwater vehicle (AUV). To address these challenges, we propose a modular underwater robot, Modularis, that can serve as an open-source testbed system. Our proposed system expedites the testing of perception, planning, and control algorithms.
\end{abstract}


\section{Introduction}

Recent years have witnessed a remarkable surge in the utilization of Autonomous Underwater Vehicles (AUVs) as a solution to the grand challenges confronting our oceans. Some examples include the realms of marine exploration, underwater mapping, environmental monitoring, underwater archaeology, and offshore industry activities. AUVs hold tremendous promise in addressing the pressing issues that demand our attention in these domains. The advent of AUVs facilitates remarkable advancements in various areas critical to underwater operations such as in mapping intricate underwater terrains, precisely localizing their positions, and devising efficient path planning strategies. These technological strides have already proven their worth in multiple applications, ranging from comprehensive surveys of unexplored marine ecosystems to high resolution documentation of underwater archaeological sites. Furthermore, AUVs prove to be instrumental in supporting offshore industry activities, optimizing resource exploration, and monitoring environmental parameters.

The future of AUV development faces three bottlenecks: cost, bring-up time, and lack of modularity. Industrial and research-focused AUVs offered by various companies are more expensive compared to aerial or ground autonomous robots of similar quality. Developing a custom AUV involves extensive mechanical design iterations, custom hardware and electronic components, and programming, often utilizing ROS1 or ROS2. Even with low-cost AUVs available in the market, inexperienced organizations may spend over a year on prototyping, posing a significant barrier to entry. Modularity is crucial for accommodating diverse end-use cases, particularly in industry and academia. Modularity in autonomous robotics pertains to flexibility in configuration across mechanical, electrical, and software designs. This paper concentrates on addressing the modularity bottleneck through the introduction of Modularis—a cost-effective, module-based AUV constructed on a BlueROV2 framework.

The modular framework introduced in our work provides contributions in AUV's mechanical, electrical, and oftware design, which enables low-cost and rapid validation of AUV systems. There are three contributions to the mechanical design built upon the BlueROV2 frame. The first contribution is the battery and storage bay, which accommodates up to two of three inch battery enclosure tubes. The second contribution is the electronics enclosure tube holder, which enables 3 different electronics enclosure tube sizes with the same frame construction. The third contribution is the stabilizer rings. Similar to the tube holder, the rings are in three sizes, one for each tube size. The rings allow the same electronics to work across all three tube sizes. For electrical design, there are two key contributions. The first contribution is the modularis main board (MMB), which is designed to accommodate plug and play sensor connections, which is enabled by the second key contribution, the sensor board. The sensor board is a small hot insertion PCB designed to work with a specific sensor, but connecting to the MMB with the same connections, regardless of sensor. This design allows for sensor boards to be designed for different applications without needing to replace the main board. The software design is developed to work specifically with the proposed hardware design, and the key contribution of the software design is the dual computer system. The dual computer system consists of two distinct bodies of software that can be used on modularis. The first contribution is a ROS1-based tethered system focused on accelerating computationally expensive experiments. The second contribution is a ROS2-based untethered system that supports a small amount of machine learning capabilities. This dual-way approach provides ease in testing of perception, planning, and control algorithms by expanding potential use cases for modularis. ROS packages for both approaches are open-source.

This paper is organized in four sections.  In Section \ref{sec:mechanical}, the mechanical design of modularis is introduced. In Section \ref{sec:electrical}, the electrical system and several circuit boards designed are presented. The schematic and PCB layout of the developed sensor board is included. In Section \ref{sec:software}, the dual computer system and ROS architecture is presented. In Section \ref{sec:experiment}, the field experiment results are presented.


\subsection{Related Work}
%
As there is a large body of research currently on AUV's, we will highlight some recent works here on which Modularis builds upon to better expound the specific contribution we make. The HAUV \cite{HAUV} was developed at UC San Diego to facilitate AUV-based vSLAM research at low-cost, developing both custom hardware and ROS packages. The authors achieved a total cost of under 13,000 USD for a high quality system built upon a BlueROV2 base. In \cite{MarketReady-AUV}, a framework is provided for effectively converting a low-cost ROV (such as BlueROV2) to an AUV. The authors focus primarily on augmenting the existing hardware and software of BlueROV2. In \cite{Frankenstein}, a more robust software architecture building upon the BlueROV2 software architecture was developed that added a handful of autonomous capabilities to BlueROV2, although in developing software on top of what BlueROV2 currently has proved to limit the extent of autonomous capabilities of the ROV. In \cite{LoCO} the authors present a low-cost AUV focused on ease of use and modularity in the physical makeup of the craft. The locations and amount of thrusters can be switched with minimal effort compared to preceding works. The software of this AUV also enables quick startup time for tasks such as person following, the power on and go capability of the LoCo AUV while being at a low-cost makes it one of the best options available. The contribution of Modularis is in extending system modularity beyond that presented in LoCo to provide an AUV of the light-weight low-cost class that can be readily available for validation of AUV experiments across a wide range of computational requirements by designing the mechanical and electrical systems to be easily extendable to three levels of computational needs, and to a large host of sensors.




\section{Mechanical Design}
\label{sec:mechanical}

\begin{figure}[t]
    \centering
    \includegraphics[width=9cm]{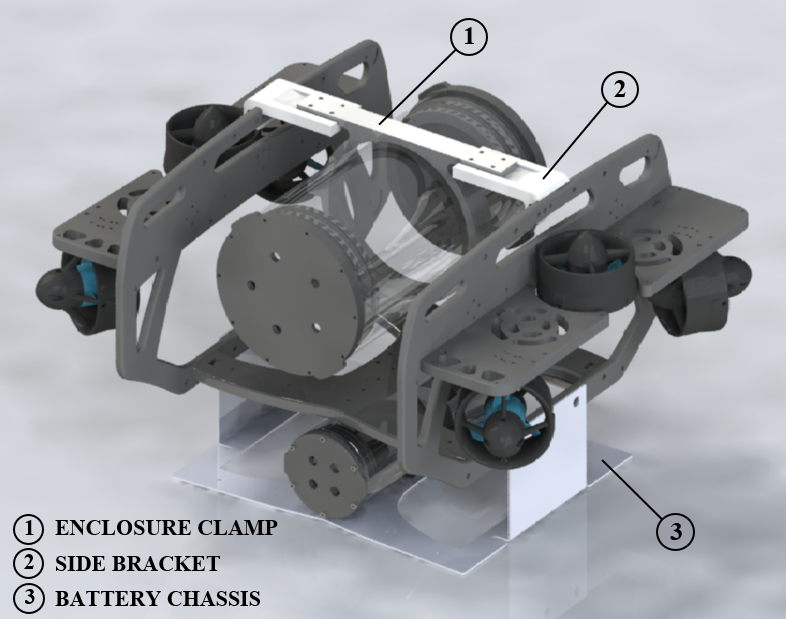}
    \caption{Assembly model of the upgraded BlueROV2.}
    \label{fig:cadmodel}
\end{figure}
\begin{figure}[t]
    \centering
    \includegraphics[width=9cm]{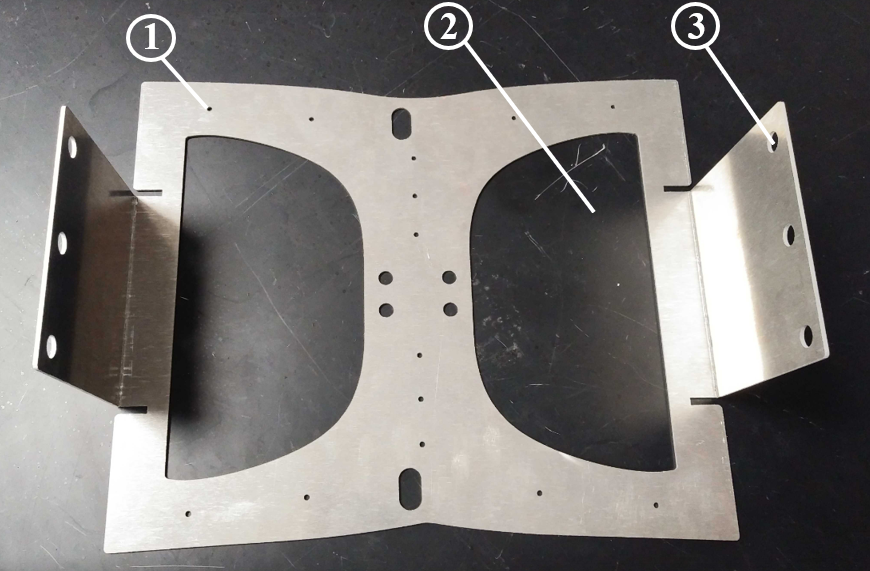}
    \caption{Sheet metal battery chassis: 1) clearance holes for ballasts, 2) cutout for weight reduction, 3) oversized mounting holes for rapid assembly.}
    \label{fig:sheetmetal}
\end{figure}
\begin{figure}[t]
    \centering
    \includegraphics[width=9cm]{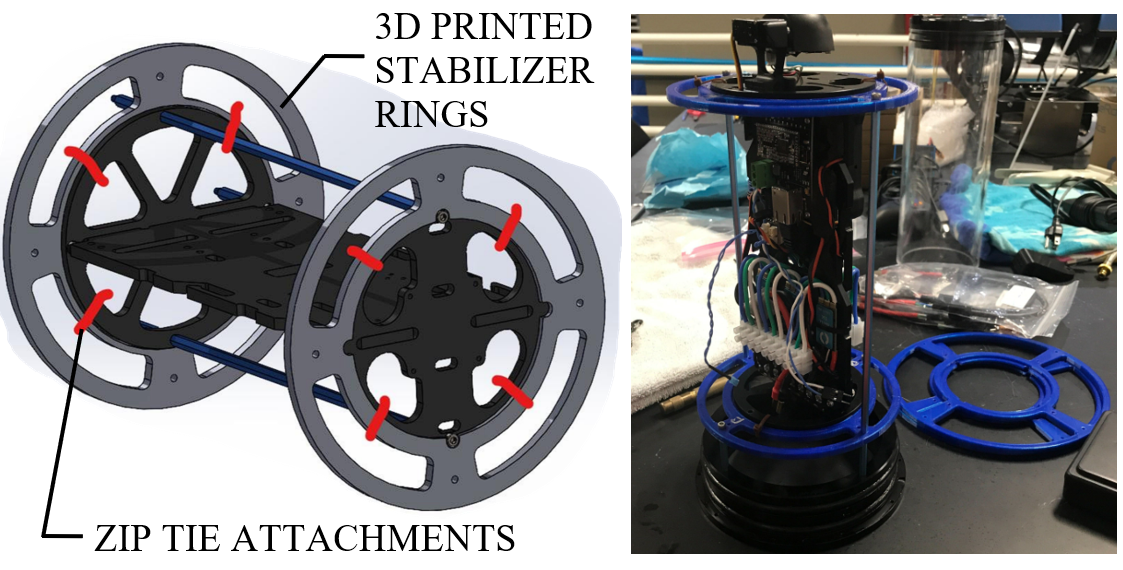}
 
    \caption{The 3D printed stabilizer rings for mounting the new electronics board inside the 6-inch diameter electronics enclosure tube.}
    \label{fig:stabilizer}
\end{figure}

The mechanical design retrofitted the BlueROV2 by Blue
Robotics into a custom AUV that can accommodate the electrical hardware necessary for the proposed autonomous design, while improving its modularity for underwater surveying operations. The 6-inch configuration is an upgraded version of the default 4-inch-diameter electronics enclosure of BlueROV2 by BlueRobotics, and has the capability to be upgraded to an 8-inch-diameter configuration using similar or scaled-up versions of preexisting parts (Fig. \ref{fig:cadmodel}). The mechanical design involved designing and fabricating a sheet metal battery chassis (Fig. \ref{fig:sheetmetal}), reconfiguring the BlueROV2 frame (Fig. \ref{fig:cadmodel}), and 3D printing parts for mounting the electronics tubes and board (Fig. \ref{fig:stabilizer}).

The battery chassis is designed from 12 gauge aluminum 5052-H32, and was waterjet cut and formed at the University of Florida. The addition of the battery chassis is necessary to relocate the battery enclosure tube to below the bottom panel of the BlueROV2, which allows larger electronics enclosure tubes to fit above the panel. The battery chassis thus expands the space available for the new electrical hardware, which otherwise does not have the space to fit inside the old 4-inch diameter enclosure. Other features of the battery chassis include clearance holes for ballast attachments, a cutout for weight reduction, and oversized mounting holes for rapid assembly onto the side panels of the BlueROV. 

The layout of the BlueROV2 is reconfigured to open up space for larger electronics enclosure tubes, which involves the relocation of the thruster panels to the outside of the frame (Fig. \ref{fig:cadmodel}). Finally, an electronics enclosure clamp and side brackets were 3D printed to mount larger electronics  tubes onto the frame of the BlueROV. Stabilizer rings were 3D printed to mount the electronics board inside the larger-diameter enclosure tubes (Fig. \ref{fig:stabilizer}).   

\section{Electrical Design}
\label{sec:electrical}

Modularis contains three main electrical sub-systems, the main, power, and sensor sub-systems. When integrated together, these sub-systems control 6 to 8 thrusters, provide power to the computer (RPi/Jetson), and interface the sensors together (IMU). Data transmission between the terminals is facilitated either through the utilization of CAN BUS Protocol or via direct connection boards using I2C/SPI. 

\subsection{Main Sub-System}
The main system PCB acts as a mounting platform for other system PCBs while providing mechanical support for electronic hardware (shown in Fig. 3), and transporting power and signals through its traces. This design choice is beneficial in replacing a hardware tray that would result in an exponential increase in the size and cost of the AUV. 

The motor control electronics are designed on the main board for the goal of space saving and efficiency. The motor control arrangement contains two 4-channel 2x1 multiplexers which provide the AUV with the ability to switch between software and hardware-controlled PWM signals. This design consideration results in the increase of software resources for image processing that would otherwise be used inefficiently.

\subsection{Power Sub-System}
The power system contains a buck converter circuit and a battery management system, housed on the same PCB.  This specific design of the buck converter circuit was utilized to regulate the voltage from a 4-cell LiPo battery to 5V which is required by multitudes of components within the AUV. Furthermore, the supply is rated to an amperage of 6 amps. Considering the safety aspect of this project, the power sub-system is designed to protect against short circuits and over-currents.

\begin{figure}[!htbp]
    \centering
    \includegraphics[width=9cm]{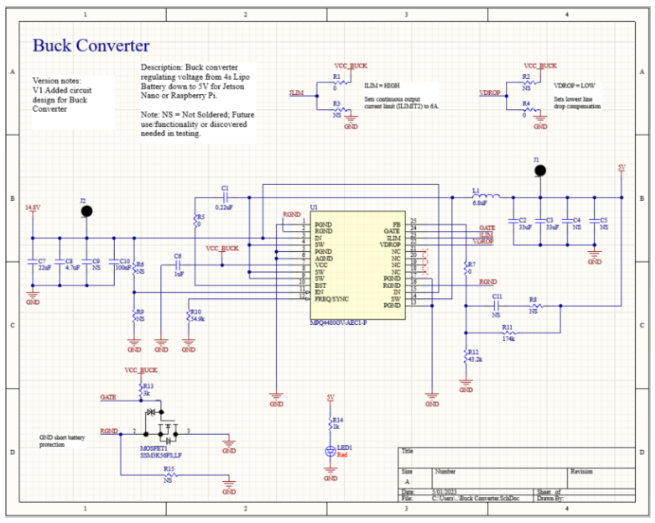}
    \caption{Buck Converter Schematic}
    \label{fig:Buck converter Schematic}
\end{figure}

Similarly to the buck converter circuit, the main safety aspect of the battery management system is to detect when fault conditions occur. These fault conditions can appear as either overvoltages or under voltages. The circuit monitors each cell of the LiPo battery using back-to-back FETs separating the supply and load. 

\begin{figure}[!htbp]
    \centering
    \includegraphics[width=9cm]{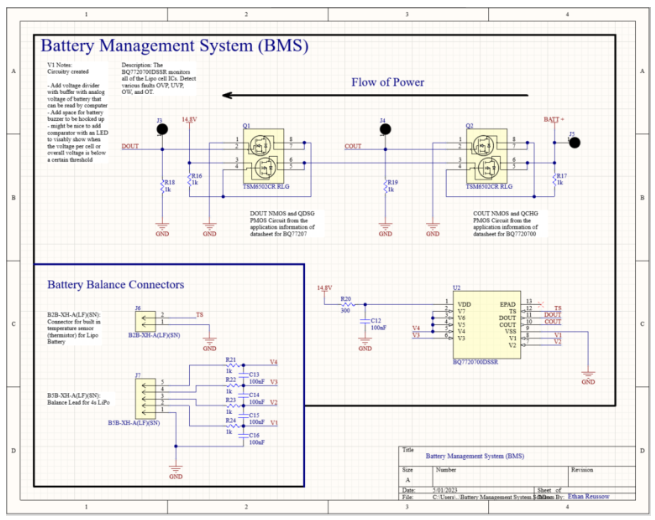}
    \caption{Battery Management System}
    \label{fig:Battery Management System}
\end{figure}

\subsection{Sensor Sub-System}
The main component of the sensor system is comprised of the BNO085 IMU board. The package was an ideal component for the system as it utilizes 9 degrees of freedom containing an accelerometer, gyroscope, and magnetometer in a single piece of hardware. Additionally, the sensor has the capabilities to provide stability detection, as well as detect when shakes and significant motion are affecting it as well.

\begin{figure}[!htbp]
    \centering
    \includegraphics[width=9cm]{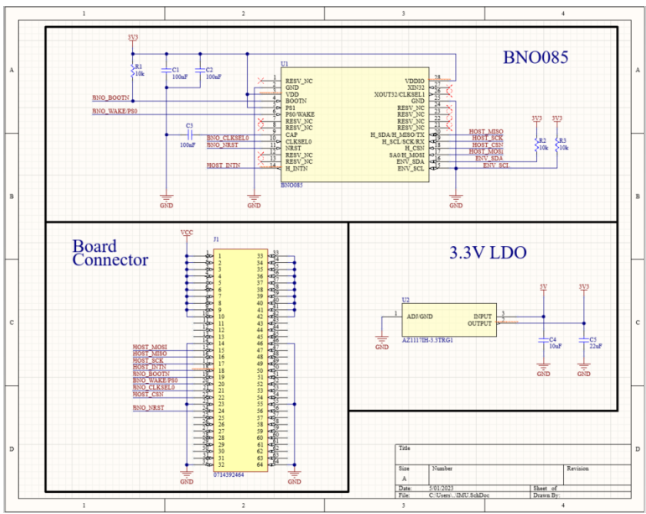}
    \caption{IMU Board}
    \label{fig:IMU Board}
\end{figure}

To ensure the digital signals are robust for various sensor ICs the CAN BUS is utilized in the AUV's system. The location of the CAN BUS situated in the sensor sub-system promotes rapid transfer of data between the utilized sensors and the computer onboard the AUV. The CAN BUS features a data rate of 1Mb/s, further enabling the high-speed transfer of data. While the Raspberry Pi has the ability to communicate with the CAN BUS, the Jetson Nano does not have any CAN hardware with this capacity. Thus, a CAN Module IC is used to communicate, via SPI, to a CAN transceiver to access the CAN BUS. 

\begin{figure}[!htbp]
    \centering
    \includegraphics[width=9cm]{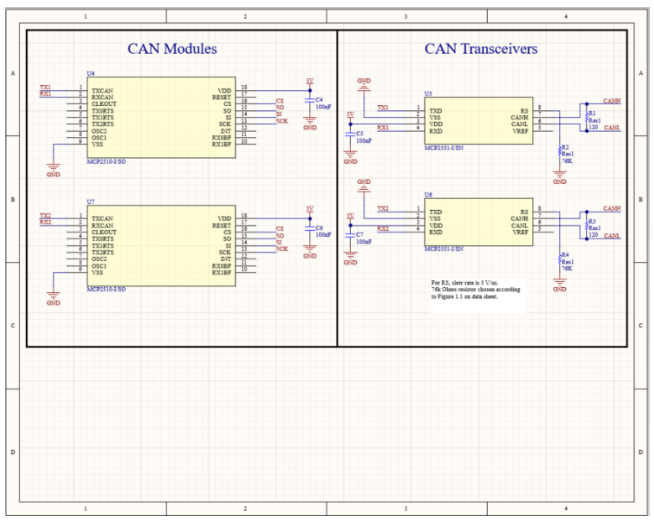}
    \caption{CAN Module and Transceiver Schematic}
    \label{fig:CAN Module and Transceiver Schematic}
\end{figure}

\begin{figure}[!htbp]
    \centering
    \includegraphics[width=9cm]{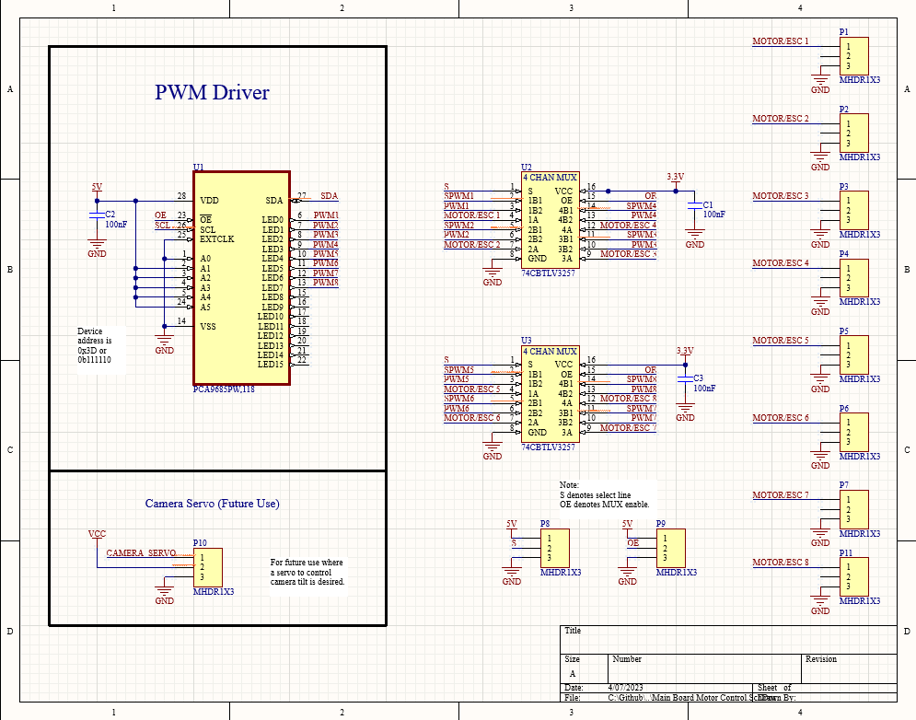}
    \caption{PCB Schematic of the motor control section using pulse width modulation (PWM) of the main system PCB}
    \label{fig:PWM}
\end{figure}




\section{Software Design}
\label{sec:software}

Modularis's software operates a dual-boot Robot Operating System (ROS) with a tethered and untethered architecture. Each system benefits research applications and adds flexibility to the AUV's operation. While tethered the AUV uses the processing power of a land-based computer, and when untethered the system relies on a Raspberry PI for all computations.

Both tethered and untethered Modularis software systems employ a modular design using ROS to easily incorporate or swap new sensors and drivers into the sensor array. This enables the AUV to test different sensor-based navigation methods depending on the task at hand. Each sensor is its own node which publishes the node-specific sensor data as a 'message'. This modular system of nodes and messages allows for quick, organized software development when adding and removing components. Currently, Modularis uses a Dell WB7022c USB camera running with an Open Robotics ROS2 USB-cam driver package which provides a configurable ROS interface to the kernel API of the libv4l2 library that implements a common driver for standard USB web cameras. With the package, we can control the camera's parameters, including frame rate, image resolution, and color format. The driver publishes the raw image data for data processing. The Modularis also uses a BNO085 IMU for position and movement data with a similar node-message configuration. Modularis is capable of utilizing not only visual data but also imu, sonar, and potentially chemical and temperature sensors.


\begin{figure}
    \centering
    \includegraphics[width=10cm]{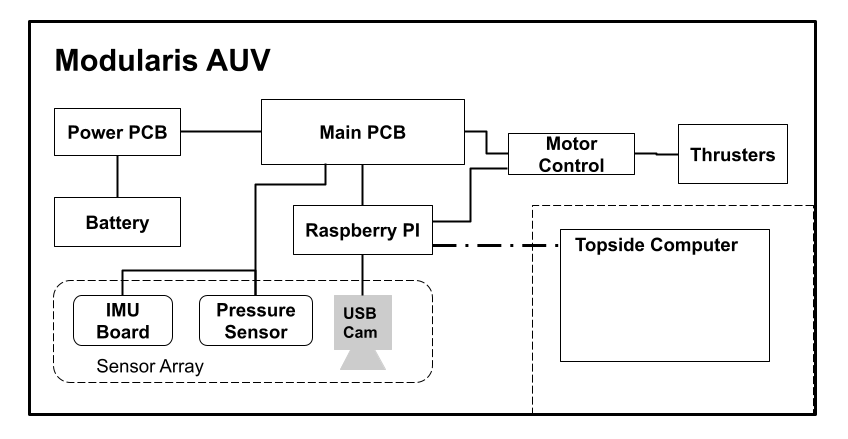}
    \caption{Diagram of Modularis's Electrical and Software Systems.}
    \label{fig:modularis_diagram_elec_soft}
\end{figure}

The thrusters are structured in a similar manner, with each one representing a unique node in ROS. This allows for a straightforward control system, making it easy for researchers to adapt it for their own needs. Each thruster is controlled by using a Blue Robotics Basic ESC (Electronic Speed Controller), which is then connected to a GPIO pin on the Raspberry PI. The ESC converts incoming PWM (Pulse Width Modulation) signals of different duty cycles from the Raspberry PI into a command that controls the thruster's speed and direction.

\begin{figure}
    \centering
    \includegraphics[width=6cm]{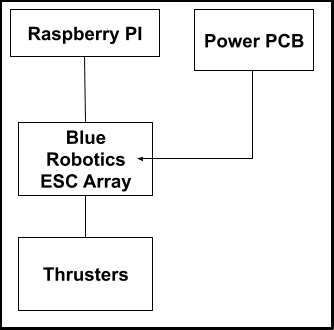}
    \caption{Simple diagram of the PWM-based Electronic Speed Controller.}
    \label{fig:speed_controller}
\end{figure}

The modularity of this system allows the AUV to run a basic onboard autonomous system with a Raspberry PI running, or connect to a tethered land-based computer allowing for more complex computations and autonomous navigation. To benefit researchers' object detection based-navigation systems, Modularis can operate with either an untethered ROS2 system or a tethered ROS1 system. Each architecture can be swapped with a reinstallation on the RassberryPI's Ubuntu operating system.

\begin{figure}
    \centering
    \includegraphics[width=8cm]{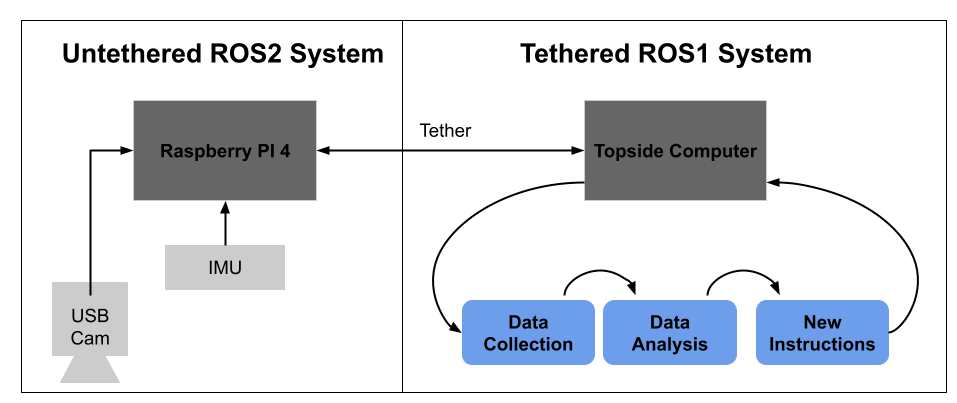}
    \caption{Diagram of the tethered and untethered systems.}
    \label{fig:thethered_untethered_diagram}
\end{figure}

\subsection{Tethered System}
When tethered, the AUV has access to more powerful land-based processors that enable a more diverse range of applications such as object detection, human recognition, and more complex autonomous navigation. The additional computational power creates a test bed for  neural-network and deep-learning research applications.

\subsection{Gazebo Simulation}
Throughout the development of the software, we used Gazebo software for simulation. Gazebo is an open-source 3D modeling and simulation software that integrates with ROS \cite{Gazebo}. The open nature and widespread adoption of the software allows for quick identification and resolution of any application bugs that may arise. Gazebo allows us to model the robot’s physical properties, such as shape and thruster locations, as well as the properties of the physical world (water, air, gravity). One example of the power of these combined features is the ability to import a CAD model of our AUV and Gazebo will simulate the force of drag caused by water. 

Gazebo also allows for simulation of the AUV’s sensors and actuators, such as camera, IMU, and thruster motors. Through it’s integration with ROS, Gazebo allows us to simulate the AUV’s code, processing sensor data and controlling thrusters as it would in deployment. This allows for more accurate insight into the AUV’s ultimate performance and enables us to identify potential situations early in the design phase. Ultimately, Gazebo provides a way to test the AUV’s software in an environment that closely mimics the actual AUV’s deployment.

\begin{figure}
    \centering
    \includegraphics[width=8cm]{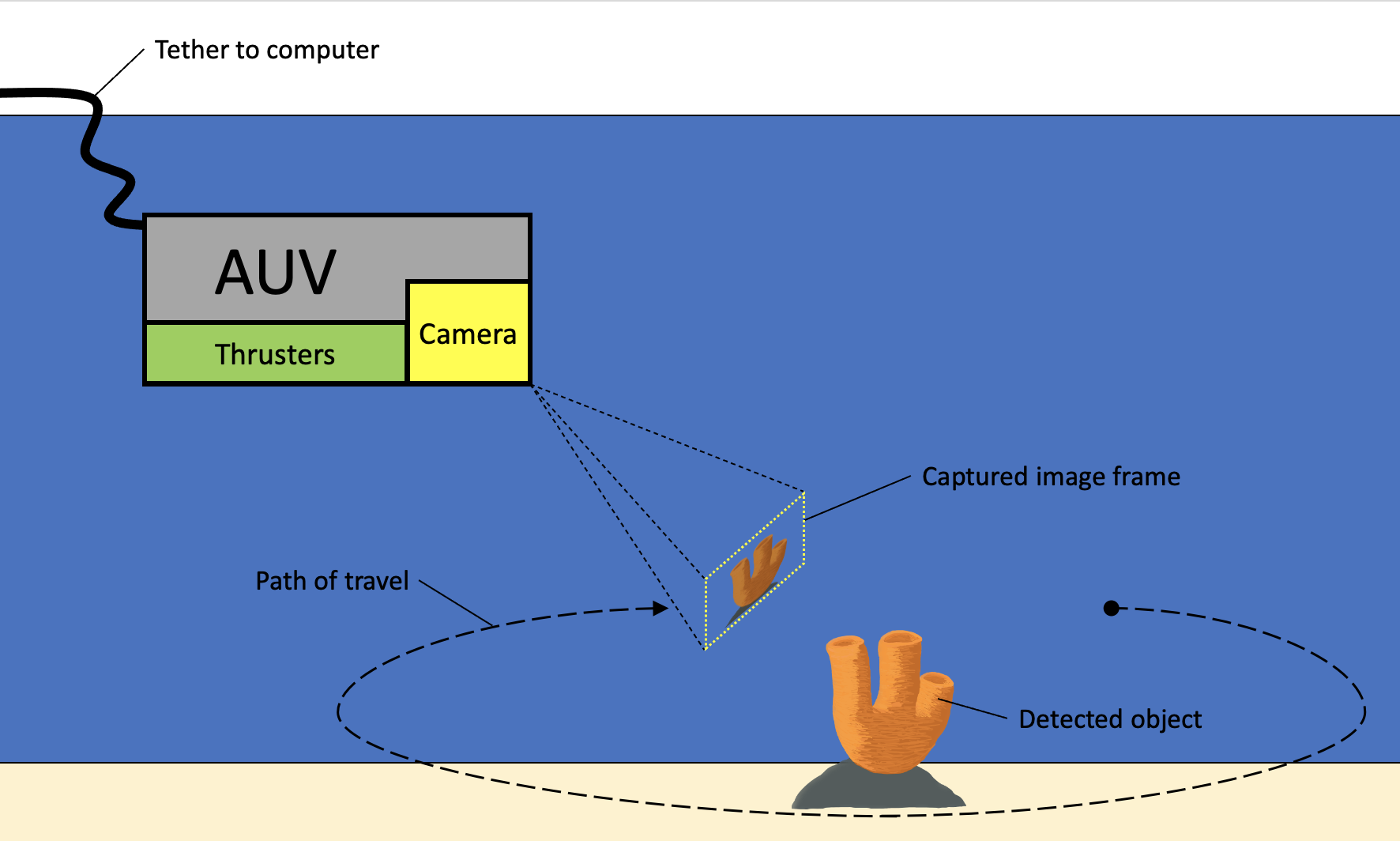}
    \caption{Diagram of how the AUV will circle a detected object and capture images while tethered to a computer.}
    \label{fig:framework_coral}
\end{figure}

Another useful feature of Gazebo was the ability to load 3D models for our simulated world. We could test our image detection software alongside the AUV's control software by providing Gazebo with the location of the cameras and thrusters on our AUV’s frame. For our image detection, we chose to use YOLOv7 \cite{YOLOv7}. This was chosen because it can be easily adopted and trained to detect new models with high accuracy. For our application, we used a ROS implementation of YOLOv7 \cite{yolo_ros}. We ran YOLOv7 on the image stream supplied by ROS, and then we were able to output the bounding box of the image. Additionally, we were able to output the AUV’s real and odometer estimated poses, i.e., position and orientation. One application of these features is the ability to simulate the process a AUV would take to build a neural radiance field. A diagram of this setup is shown in Fig. \ref{fig:framework_coral}.

Although the Gazebo environment allows us to simulate the AUV’s environment and physical properties, the simulation did not necessarily provide an accurate representation of the final AUV’s computer hardware. For example, while the gazebo simulation worked fine for simulating the AUV on a laptop, the overhead was too much for our Jetson Nano to process. On the other hand, running the simulation in this way also allows us to identify potential limitations in the Jetson Nano and opened a discussion as to whether our system needed more powerful computing elements.

Unfortunately, the single camera setup with YOLOv7 allows only for object detection in the frame. For advanced applications, we ultimately will need the object’s estimated location relative to the AUV. For this there are several options to consider. One of which is 3d stereo imaging, which would require the introduction of a second camera. One benefit of this is that it is simple to implement. Limitations would be the space required in the AUV body for a second camera. additionally, stereo imaging requires the AUV to remain relatively motionless while capturing the images. Another option would be to implement a mixed imaging system, that would combine the images with other sensor data, such as sonar or lidar, to create a 3D model of the AUV’s perceived environment. Another potential solution would be to build a neural radiance field from a sequence of 2D images taken at multiple points around the object.

\section{Experimental Results}

\label{sec:experiment}

\subsection{Buoyancy Test}
When changing the electronic enclosure tube, the buoyancy changes dramatically. To keep the sub neutrally buoyant, the weight needs to be easily adjusted. Our design, using dive weight belts looped around the frame of the sub, allows for the simple weight management. Through field tests, we found that adding 4lbs to the 4.5in and 15lbs to the 6in enclosure configuration keeps the sub neutrally buoyant in chlorinated water. Prematurely finding these values has reduced the deployment time. Future experiments will provide more information on how much added weight will be needed for varying water salinities.

\begin{figure}[hbt!]
    \centering
    \includegraphics[width=6cm]{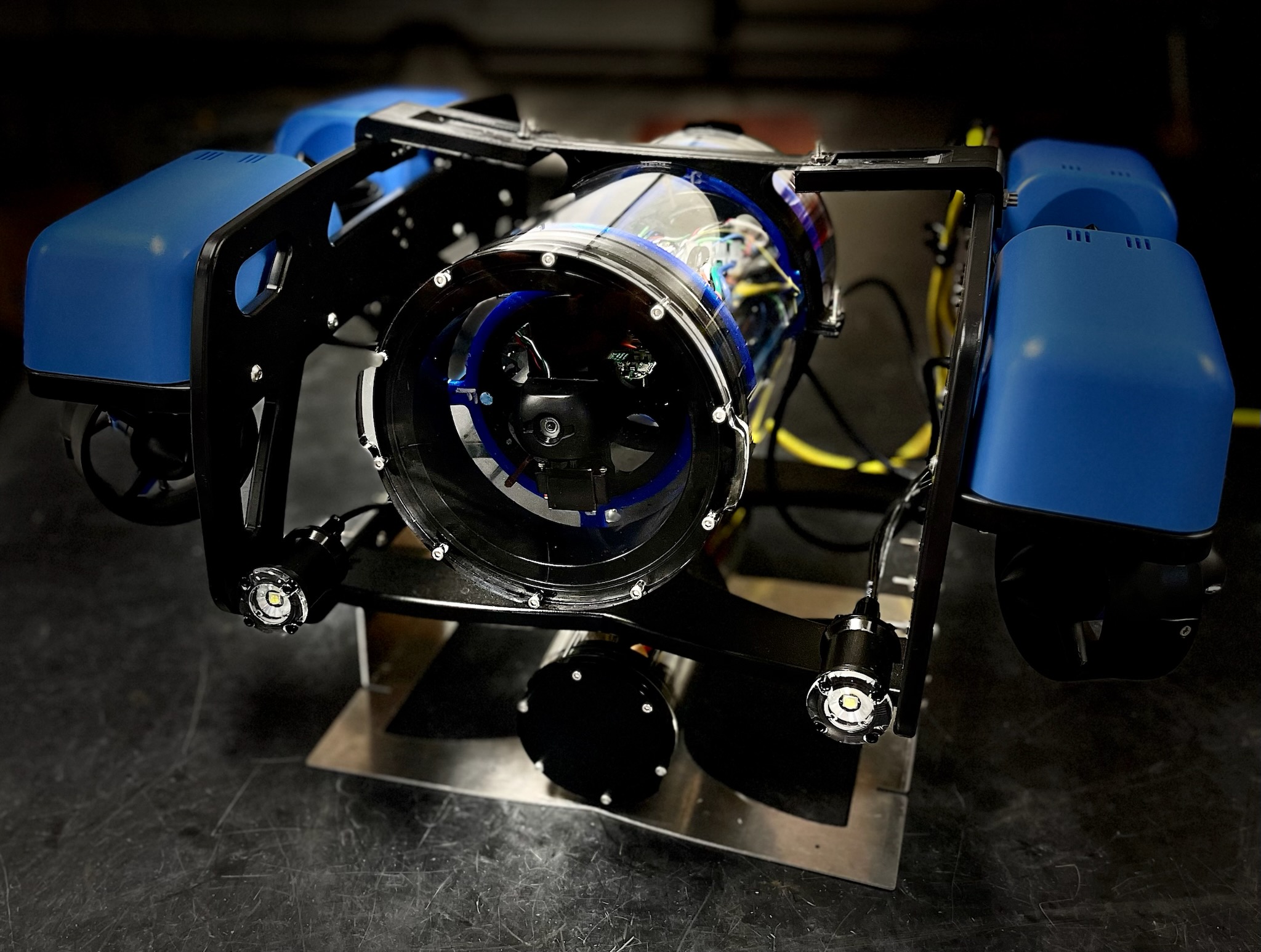}
    \caption{Fully assembled AUV with mechanical augmentations.}
    \label{fig:AUV}
\end{figure}

\subsection{Electronics Testing}

The electronics boards has been assembled and functioned correctly according to design. We plan on conducting more rigorous testing to validate the modularity of the electrical hardware across several use cases in future work. For future work we plan to test our electrical hardware and two software systems by implementing person following on Modularis in two cases- one where the AUV is autonomously controlled from a computer over tether, and another where control is tetherless, controlled solely from the onboard computer. This experiment will be conducted both in our indoor testing facility and in the field.

\begin{figure}[hbt!]
    \centering
    \includegraphics[width=6cm]{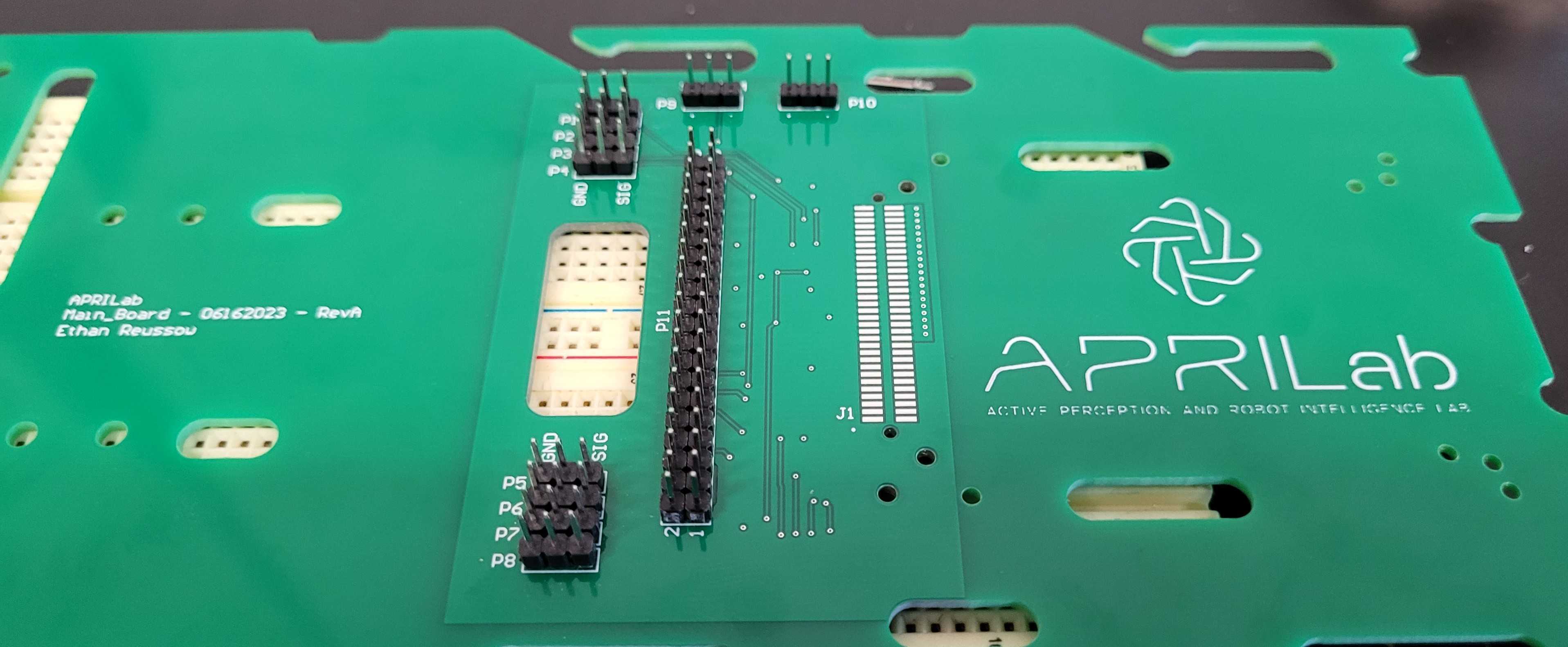}
    \caption{Main Modularis PCB board, which serves as a base for connecting processor, sensors and thrusters.}
    \label{fig:AUV}
\end{figure}


\section{Conclusions}

This paper presents an AUV design for use by industry and researchers for rapidly validating localization, planning, and perception algorithm implementations. Applications span from environmental monitoring to archaeological site mapping. Modularis has demonstrated its capability for conducting autonomous missions both un-tethered and tethered in both controlled and uncontrolled environments in the experiments presented. In the future the Modularis design will be iterated upon as greater intuition is gained from field experiments such as that conducted in Ginnie Springs. Being open-source in code and hardware design, the authors look forward to seeing improvements to software and hardware modularity, as well as exciting applications taken on by other researchers and AUV enthusiasts.

\section*{Acknowledgment}
The authors would like to thank the support from the BlueROV community, as well as University of Florida's Machine Intelligence Lab (MIL) for providing guidance in understanding the BlueROV system and providing support with the SubjuGator code base respectively. The authors also would like to acknowledge Alex Baker's help on soldering work.

\bibliographystyle{IEEEtran}
\bibliography{main}

\end{document}